\documentclass[runningheads]{llncs}
\usepackage{graphicx}
\usepackage{tikz}
\usepackage{comment}
\usepackage{amsmath,amssymb} % define this before the line numbering.
\usepackage{color}
\usepackage{booktabs}
\usepackage{makecell}

\usepackage[
bookmarksdepth=3, %bookmark levels in the PDF.
bookmarksnumbered, %Show section numbering in bookmark
colorlinks=true,
linkcolor=red, 
citecolor=teal]{hyperref}

\newcommand{\etal}{\textit{et al.}}
\newcommand{\eg}{\textit{e.g.,}}
\newcommand{\ie}{\textit{i.e., }}

\usepackage{multirow}

\begin{document}
% \linenumbers
\pagestyle{headings}
\mainmatter
\def\ECCVSubNumber{20}  % Insert your submission number here

%\title{TBSC-Net: End-to-End Trainable 3D Textured Body Shape Completion Network} % Replace with your title
\title{TSCom-Net: Coarse-to-Fine 3D Textured Shape Completion Network}
\begin{comment}
\titlerunning{ECCV-22 submission ID \ECCVSubNumber} 
\authorrunning{ECCV-22 submission ID \ECCVSubNumber} 
\author{Anonymous ECCV submission}
\institute{Paper ID \ECCVSubNumber}
\end{comment}
%******************

%\begin{comment}
\titlerunning{TSCom-Net}
\author{Ahmet Serdar Karadeniz \and
Sk Aziz Ali \and
Anis Kacem \and
Elona Dupont \and
Djamila Aouada} 
\authorrunning{Karadeniz et al.}
\institute{SnT, University of Luxembourg, L-1511 Luxembourg
\\
\email{\{firstname.lastname\}@uni.lu}}
%\end{comment}
%******************
\maketitle

\begin{abstract}
Reconstructing 3D human body shapes from 3D partial textured scans remains a fundamental task for many computer vision and graphics applications -- \textit{e.g.}, body animation, and virtual dressing. We propose a new neural network architecture for 3D body shape and high-resolution texture completion -- TSCom-Net -- that can reconstruct the full geometry from mid-level to high-level partial input scans. We decompose the overall reconstruction task into two stages -- first, a joint implicit learning network (SCom-Net and TCom-Net) that takes a voxelized scan and its occupancy grid as input to reconstruct the full body shape and predict vertex textures. Second, a high-resolution texture completion network, that utilizes the predicted coarse vertex textures to inpaint the missing parts of the partial \lq texture atlas\rq. A Thorough experimental evaluation on 3DBodyTex.V2 dataset shows that our method achieves competitive results with respect to the state-of-the-art while generalizing to different types and levels of partial shapes. The proposed method has also ranked second in the track1 of SHApe Recovery from Partial textured 3D scans (SHARP \cite{saint2020sharp,SHARP2022Repo}) 2022\footnote{\url{https://cvi2.uni.lu/sharp2022/}} challenge1.
%and also first with \textbf{0.5??}$\%$ of texture score when a better shape prediction is used.     

% SCom-Net Second, we employ a joint 
% fusion strategy inside SCom-Net to combine the advantages of both the geometry-based and texture-based deep features in high-quality shape reconstruction. 

% We further intro-
% duce a part-style loss to help reconstruct high-fidelity col-
% ors without introducing unpleasant artifacts. 

\end{abstract}

\section{Introduction}
\begin{figure}[!ht]
    \centering
	\includegraphics[width=0.85\textwidth]{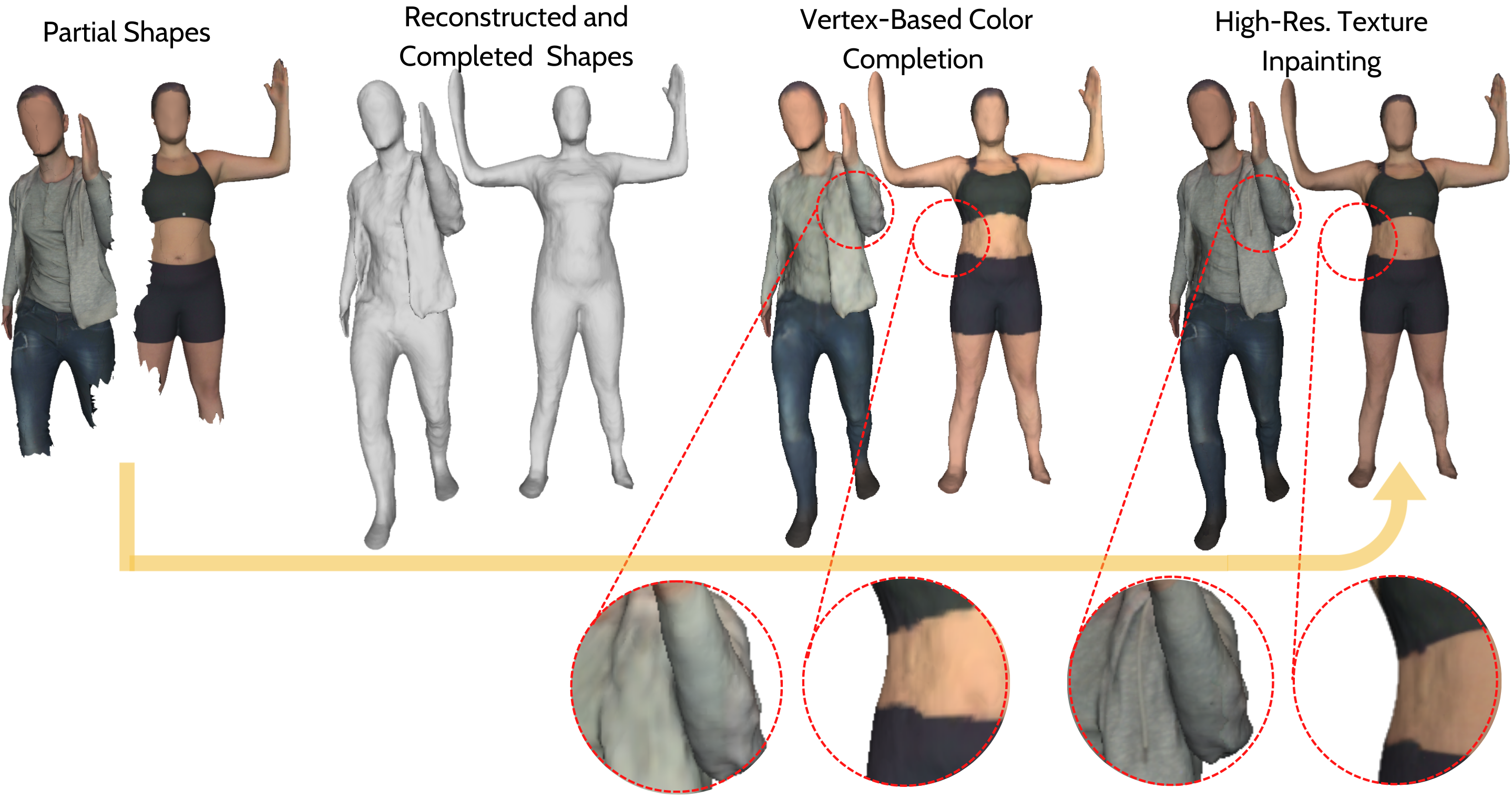}
    \caption{\textbf{3D Partial Textured Body Shape Completion}. The proposed deep learning-based method reconstructs and completes the surface geometry of a dressed or minimally-clothed partial body scan, and inpaints the missing regions with high-resolution texture. TSCom-Net has the flexibility to either apply dense texture mapping or vertex-based color estimation as per desired application scenarios. 
    }
    \label{fig:Bcomnet_TEASER}
\end{figure}
3D textured body shape completion and reconstruction from visual sensors plays a key role in a wide range of applications such as gaming, fashion, and virtual reality~\cite{prokudin2021smplpix,patel2021agora}. The challenges of this task and the methods to tackle it varies depending on the used sensors to capture the human shape and texture. Some existing methods focus mainly on 3D textured body shape reconstruction from a single monocular image~\cite{AngjooECCV18,wangCVPR19,liECCV20,hasler2010multilinear,zheng2019deephuman}.
%or multi-view 2D images \cite{liECCV20}. 
For instance, under the body shape symmetry assumption, methods like Human Mesh Recovery (HMR)~\cite{AngjooCVPR18}, use a statistical body model~\cite{SMPLTOG15} to reconstruct the body shape and pose. Despite the impressive advances in this line of work~\cite{wangCVPR19,liECCV20,zheng2019deephuman}, the reconstructed shapes still cannot capture high-level geometrical details, such as wrinkles, ears. This is due to the lack of original geometrical information and the projective distortions in 2D images. On the other hand, 3D scanners and RGB-D sensors can provide richer information about the geometry of body shapes~\cite{tian2022recovering}. Nevertheless, they are often subject to partial acquisitions due to self occlusions, restricted sensing range, and other limitations of scanning systems~\cite{yan2022shapeformer,chibane20ifnet}. While most of existing works focus on completing the geometry of body shapes from 3D partial scans~\cite{chibane20ifnet,yan2022shapeformer,sarkar2017learning}, less interest has been dedicated to complete both the texture and the geometry at the same time. Nonetheless, this problem remains critical in real-world applications where complete and realistic human reconstructions are usually required.
Aware of this need, recent competitions, such as SHApe Recovery from Partial Textured 3D Scans (SHARP) challenges~\cite{SHARP2022,SHARP2021,saint2020sharp}, emerged in the research community to foster the development of joint shape and texture completion techniques from partial 3D acquisitions. The results of these competitions showed promising techniques~\cite{chibane2020ifnet_texture,saint20203dbooster} with a room for improvements. 

\vspace{0.1cm}
The problem of textured 3D body shape completion from 3D partial scans, as shown in Fig.~\ref{fig:Bcomnet_TEASER}, can naturally be decomposed into two challenging subtopics -- (i) partial shape surface completion and (ii) partial texture completion of the reconstructed shape.
%This problem, as shown in Fig.~\ref{fig:Bcomnet_TEASER}, imposes different challenges than the aforementioned image-based approaches. 
In this setup, especially when no fixed UV-parametrization of shape is available, methods need to rely on inherent shape structure and texture correlation cues from the available body parts. 
%This partial textured shape completion problem is more alike with some recent works~\cite{chibane2020ifnet_texture,saint20203dbooster} and closely with ShapeFormer~\cite{yan2022shapeformer} (suitable only for shape completion tasks). 
Taking this direction, IFNet-Texture~\cite{chibane2020ifnet_texture} uses implicit functions to perform high-quality shape completion and vertex-based texture predictions. Despite the impressive results in SHARP challenges~\cite{saint2020sharp,SHARP2021}, the method~\cite{chibane2020ifnet_texture} does not output high-resolution texture and often over-smooths vertex colors over the whole shape. On the other hand, 3DBooSTeR method~\cite{saint20203dbooster} decouples the problems of shape and texture completions and solves them in a sequential model. The shape completion of~\cite{saint20203dbooster} is based on body models~\cite{groueix20183d}, while the texture completion is tackled as an image based \textit{texture-atlas} inpainting. \cite{saint20203dbooster} can complete high-resolution partial texture, but suffers from large shape and texture modelling artifacts. 
%Halso recovers a complete mesh from partial scans and later inpaint high-resolution partial texture but suffers from large shape and texture modelling artifacts.   

\vspace{0.1cm}
Our method overcomes the weaknesses of both \cite{chibane2020ifnet_texture} and \cite{saint20203dbooster}. The shape completion part of our TSCom-Net is designed to learn a continuous \textit{implicit function} representation~\cite{mescheder2019occupancy,Zheng_2021_CVPR,park2019deepsdf} as also followed by Chibane~\etal~\cite{chibane20ifnet}. Furthermore, we identify that the IFNet-Texture method~\cite{chibane2020ifnet_texture}, an extension of~\cite{chibane20ifnet}, uses a vertex-based inference-time texture estimation process that ignores the shape and color feature correlation cues at the learning phase. 
For this reason, we employ an end-to-end trainable vertex-based shape and texture completion networks -- SCom-Net and TCom-Net -- to boost their performance with joint-learning. 
%This is the first layer of improvements of TSCom-Net.
Next, we propose to refine the predicted vertex-based texture completions directly in the texture-atlas of 3D partial scans. This is achieved using an image inpainting network~\cite{liu2018partialinpainting,saint20203dbooster}, reusing the predicted vertex textures, and yielding high-resolution texture.
It is notable that, unlike~\cite{lazova3DV19,saint20203dbooster}, TSCom-Net does not require a fixed UV-parametrization which makes the considered \lq texture charts\rq~random, discontinuous, and more complex in nature across the samples. Overall, our \textbf{contributions} can be summarized to: 
\begin{enumerate}
    \item An end-to-end trainable joint implicit function network for 3D partial shape and vertex texture completion. We propose an early fusion scheme of shape and texture features within this network.
    \item A high-resolution texture-atlas inpainting network, which uses partial convolutions~\cite{liu2018partialinpainting} to refine the predicted vertex textures. At the same time, this module is flexible and can be plugged with any other 3D shape and vertex texture reconstruction module. 
    \item An extensive experimental evaluation of TSCom-Net showing its multi-stage improvements, its comparison \textit{w.r.t.} the participants in SHARP 2022 challenge~\cite{SHARP2022}, and its generalization capabilities to different types of input data partiality. 
\end{enumerate}

\vspace{0.1cm}
The rest of the paper is organized as follows. After summarizing the methods related to our line of work in Section~\ref{sec:related_works}, we present the core components of TSCom-Net in Section~\ref{sec:ProposedApproach_TSCom-Net}. The experiments and evaluation parts are reported and discussed in Section~\ref{sec:Experiments_Results_Evaluation}. Finally, Section~\ref{sec:Conclusion} concludes the paper and draws some future works. 

\vspace{-0.2cm}
\section{Related Works}\label{sec:related_works}
\vspace{-0.15cm}
\noindent\textbf{Deep Implicit Function and 3D Shape Representation Learning. } Supervised learning methods for 3D shape processing % -- for instance, reconstruction, completion, alignment, synthesis and compression -- 
require an efficient input data representation. Apart from common learning representations of 3D shapes -- such as regular voxel grids~\cite{liu2019point,zheng2019deephuman}, point clouds \cite{qi2017pointnet}, Oc-trees~\cite{wang2017cnn}, Barnes-Hut $2^D$-tree~\cite{ali2021rpsrnet}, depth-maps~\cite{malik2020handvoxnet}, and meshes~\cite{hanocka2019meshcnn}, the popularity of implicit representation~\cite{mescheder2019occupancy,Zheng_2021_CVPR,park2019deepsdf} has recently increased.   
%Deep implicit functions (either of signed distance \cite{Zheng_2021_CVPR,park2019deepsdf} or truncated signed distance fields \cite{zheng2019deephuman}) 
These representations~\cite{Zheng_2021_CVPR,park2019deepsdf} serve as a continuous representation on the volumetric occupancy grid~\cite{mescheder2019occupancy} to encode the iso-surface of a shape. Therefore, given any query 3D point, it returns its binary occupancy in the encoded iso-surface. This continuous representation is extremely useful when the input shapes are partial and the spatial locations can be queried on missing regions based on the available regions.

%\vspace{0.1cm}
%Deep implicit functions (either of signed distance \cite{Zheng_2021_CVPR,park2019deepsdf} or truncated signed distance fields \cite{zheng2019deephuman}) serve as a continuous representation on the volumetric occupancy grid \cite{mescheder2019occupancy} to encode the iso-surface of a shape. Therefore, given any query point coordinate, it returns the nearest signed distance from the encoded iso-surface. This continuous representation is extremely useful when the input shapes are partial and the spatial locations can be queried on missing regions based on the available regions.

\vspace{0.1cm}
\noindent\textbf{Learning-based 3D Shape Completion.}
3D reconstruction and completion of shapes, especially for human 3D body scans, is tackled in different ways~\cite{han2019image,wang2018pixel2mesh,tatarchenko2017octree,sinha2017surfnet,han2017high,yan2022shapeformer,yuan2018pcn,gurumurthy2019high}. Some methods deal with partial point clouds as input~\cite{han2017high,yan2022shapeformer,yuan2018pcn,gurumurthy2019high}. A subset of these methods do not consider textured scans and focus on different types of shape quantization -- \eg~voxel grids~\cite{han2017high}, octrees, and sparse vectors~\cite{yan2022shapeformer} with the aim of capturing only fine geometric details. Furthermore, many of these models~\cite{yan2022shapeformer,arora2022multimodal,gurumurthy2019high} cannot always predict a single shape corresponding to the partial input shape. Another traditional way for body shape completion from a set of partial 3D scans is by non-rigidly fitting a template mesh~\cite{ali2018nrga,golyanik2017framework,ali2020foldmatch}. However, improper scaling~\cite{ali2018nrga}, computational speed~\cite{ali2020foldmatch}, and articulated body pose matching~\cite{golyanik2017framework} is a major problem for these methods.

%Interestingly, Sarkar~\etal~\cite{sarkar2017learning} show that dictionary learning on \textit{quadrangulated} patches can fill small random holes in scans. \textcolor{red}{However, it is not certain that these methods can cope with the different levels of partiality found in SHARP-2022~\cite{SHARP2022} dataset}.
%\vspace{0.1cm}

%Moreover, the final shape will require additional texture completion with fixed UV-parametrization of template mesh.

\vspace{0.1cm}
More closely related works to ours are ~\cite{chibane2020ifnet_texture,saint20203dbooster,chibane20ifnet}. Saint \etal~\cite{saint20203dbooster} and Chibane \etal~\cite{chibane2020ifnet_texture} solve the same 3D textured shape completion, where the former uses 3D-Coded~\cite{groueix20183d} and non-rigid refinement to reconstruct the shape and the latter uses deep implicit functions~\cite{chibane20ifnet}. While~\cite{saint20203dbooster} provides a fixed UV-parametrization useful for the texture completion, it cannot recover extreme partial body scans due to its restriction to a template body model~\cite{SMPLTOG15}. On the other hand,~\cite{chibane20ifnet} can recover corase-to-fine geometrical details using multiscale implicit representations. However, the implicit representations cannot preserve the UV-parametrization of the input partial shape, which restricts the texture completion to a vertex-based solution~\cite{chibane2020ifnet_texture}. TSCom-Net builds on top the implicit shape representation of~\cite{chibane20ifnet,chibane2020ifnet_texture} and extends it to produce higher-resolution texture. 

%\vspace{0.1cm}
%3D reconstruction and completion of shapes \cite{han2019image,wang2018pixel2mesh,tatarchenko2017octree,sinha2017surfnet}, especially for human 3D body scans, is tackled in different ways -- \eg~a number of methods~\cite{arora2022multimodal,wu2020multimodal,han2017high,yan2022shapeformer,yuan2018pcn,gurumurthy2019high} deal with partial point clouds as input. A subset of these do not tackle textured scans and focus different types shape quantization -- \eg~voxel grids~\cite{han2017high}, octrees, sparse vectors~\cite{yan2022shapeformer} for only capturing fine geometric details. Furthermore, there exists a common problem that many of these models ~\cite{yan2022shapeformer,arora2022multimodal,gurumurthy2019high} cannot always predict a single shape corresponding to the partial input shapes. 

% In other words, their predictions are multi-modal because there is an inherent shape and structure ambiguity for generic objects.

%\vspace{0.1cm}
%RGB image-based body shape reconstruction or completion is another line of work~\cite{AngjooCVPR18,omran2018neural,bogo2016keep,hasler2010multilinear,zheng2019deephuman}. These methods are not directly related to the underlying problem statement of ours as we have unique 3D shape artifacts to correct directly in 3D, and inpaint directly in texture atlas instead of images. 
% At the same time, we are avoiding shape prior~\cite{SMPLTOG15} like other approaches~\cite{AngjooCVPR18,omran2018neural,bogo2016keep,hasler2010multilinear,zheng2019deephuman}. 

%
% On quadtringulated patches
%

\vspace{0.1cm}
\noindent\textbf{Learning-based 3D Textured Shape Completion. } 
%
% must include texturify, 3DBooster IFNet Texture, UNet and 3D-Partial Convolution-based Inpainting 
%
%
% Some methods that complete texture based on Fixed UV Map, 
%
Completing the texture of 3D body shapes is a challenging task due to many factors -- \eg~varying shades, unknown clothing boundaries, and complex styles and stripes. This task can be either tackled by completing the RGB texture for each vertex of the completed mesh~\cite{chibane2020ifnet_texture}, or directly completing the texture-atlas~\cite{saint20203dbooster,deng2018uv}. While the former takes advantage of the shape structure of the vertices to predict textures, its resolution depends on the resolution of the shape, which is often limited due to memory constraints. On the other hand, the texture-atlas completion allows for high-resolution texture generation, but does not take advantage of the structure of the shape when no fixed UV-parametrization is given~\cite{lazova3DV19,saint20203dbooster}, or multiple charts are provided~\cite{saint20203dbooster,deng2018uv}.  To adress these limitations, TSCom-Net uses both paradigms. Firstly, it completes the vertex-based textures, then refines them in the texture-atlas image space using a dedicated inpainting module. 

\section{Proposed Approach -- TSCom-Net}\label{sec:ProposedApproach_TSCom-Net}
\vspace{-0.15cm}
\begin{figure}[!ht]
    \centering
	\includegraphics[width=0.99\textwidth]{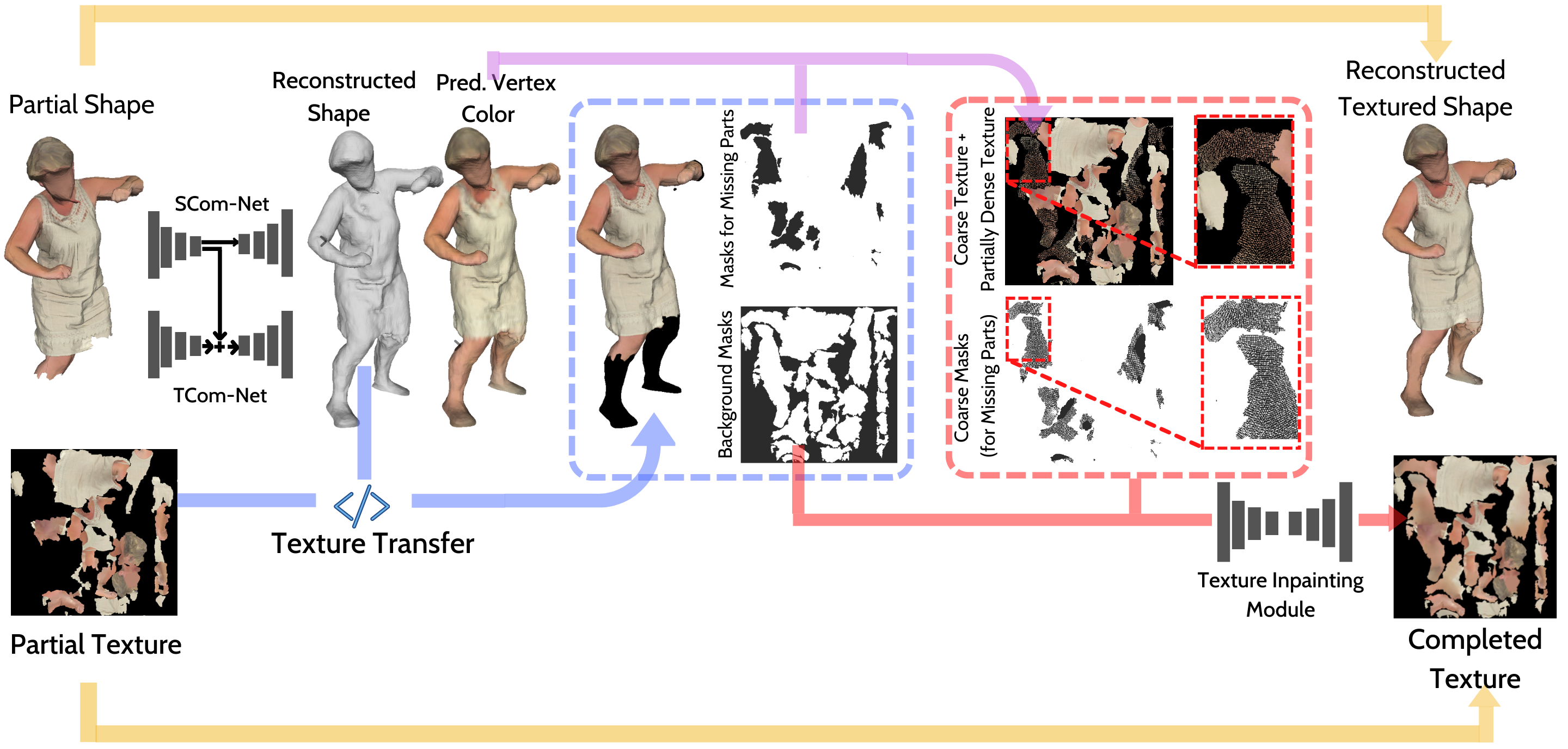}
    \caption{\textbf{Overview of TSCom-Net.} TSCom-Net has two main components -- (I) an end-to-end trainable vertex-based shape and color completion network  (\textit{on the left}) that returns low resolution and often over-smoothed vertex colors. The underlying architecture of this modules is based on IFNet \cite{chibane20ifnet} and IFNet-Texture \cite{chibane2020ifnet_texture} along with newly added modifications to boost both the shape and color completion jointly. (II) A coarse-to-fine texture inpainting module (using 2D partial convolutions \cite{liu2018partialinpainting}) that takes partially textured and fully reconstructed body shape, background coarse masks, and coarse-texture and its masks for the missing regions as input, and outputs a high-resolution, complete \lq texture atlas\rq.}
    \label{fig:Bcomnet_Overview}
\end{figure}

Given a 3D partial body scan $\mathbf{X}=(\mathbf{\mathcal{V},\mathcal{T}, \mathcal{E}})$, tuple defining set of vertices $\mathcal{V}$, triangles $\mathcal{T}$, and edges $\mathcal{E}$ along with its corresponding partial texture $\mathbf{\mathcal{A}}$, our aim is to recover the 3D complete body scan $\mathbf{Y}$ with high-resolution texture. To achieve that, we propose a deep learning-based framework TSCom-Net with two intermediate stages (see Fig.~\ref{fig:Bcomnet_Overview}). First, we reconstruct the shape and coarse vertex texture of the given partial scan. Then, we further inpaint the partial texture atlas with another network in a coarse-to-fine manner.

\subsection{Joint Shape and Texture Completion}\label{subsec:JointShapeTex_Learning}
% Purpose of two networks.

Let $\mathbf{\mathcal{X}}_s = \mathbb{R}^{N\times N \times N}$ be the discretized voxel grid of the input partial scan $\mathbf{X}$ with resolution $N$. Each occupied voxel has the value of $1$ ($0$ otherwise). Both the ground-truth and partial scans are normalized. Similarly, another voxel grid, $\mathbf{\mathcal{X}}_c$, is constructed with the texture of the input partial scan $\mathbf{X}$. 
%which is a colored voxelization of the same scan $\mathbf{X}$. 
Each occupied voxel of $\mathbf{\mathcal{X}}_c$ has an RGB texture value in $\mathbf{[0,1]}^3$ obtained from the partial texture ($\langle-1,-1,-1\rangle$ otherwise).

%\vspace{0.1cm}
TSCom-Net consists of two implicit feature networks SCom-Net and TCom-Net that learn to complete the shape and texture of a partial surface, respectively. Different from \cite{chibane2020ifnet_texture}, the two sub-networks are trained jointly to enable the interaction between shape and texture learning. An early fusion technique is utilized to improve the color accuracy by incorporating the shape features in addition to the color features of the partial scan. Consequently, the texture network TCom-Net learns to use the shape information extracted by SCom-Net to predict a more accurate texture. In what follows, we start by describing the shape completion, then present the texture completion. 
%In what follows, the shape encoding and decoding are firstly described, then the texture encoding 
%for a given point.

\vspace{0.1cm}
\noindent \textbf{Shape Completion:} The input discretized voxel grid $\mathbf{\mathcal{X}}_s$ is encoded as in~\cite{chibane20ifnet} using a 3D convolutional encoder $g_s$ to obtain a set of $n$ multiscale shape features,
\vspace{-0.15cm}
\begin{equation}
g_s(\mathbf{\mathcal{X}}_s) := \mathbf{S}_1, \mathbf{S}_2, \dots, \mathbf{S}_n  \ ,
\end{equation}
where $\mathbf{S}_i \in \mathbb{R}^{d_i\times K \times K \times K}$, $1 \leq i \leq n$,  denote the shape features with resolution $K=\frac{N}{2^{i-1}}$ and channel dimension $d_i=d_1 \times 2^{i-1}$. Accordingly, $\mathbf{S}_n$ would have the lowest resolution but the highest channel dimensionality among other shape features. Given the features $\mathbf{S}_i$, the completion of the shape is achieved by predicting the occupancies of the query points $\{\mathbf{p}_j\}_{j=1}^M$, where $\mathbf{p}_j \in \mathbb{R}^3$. At training time, the points $\mathbf{p}_j$ are sampled from the ground-truth $\mathbf{Y}$. During inference time, $\mathbf{p}_j$ are the centroids of all voxels in $\mathcal{X}_s$. Multiscale features $\mathbf{S}_{i,j}^p := \phi(\mathbf{S}_i ,\mathbf{p}_j)$ can be extracted for each point $\mathbf{p}_j$ using a grid sampling function $\phi$, taking and flattening the features of $\mathbf{S}_i$ around the neighborhood of $\mathbf{p}_j$.

%Let $\phi(\mathbf{S}_i ,\mathbf{p}_j)$ denote the grid sampling function, taking and flattening the features of $\mathbf{S}_i$ around the neighborhood of $\mathbf{p}_j$. 

%\vspace{0.15cm}
Finally, $\mathbf{p}_j$ and $\mathbf{S}_{i,j}^p$ are decoded with $f_s$ consisting of sequential 1D convolutional layers to predict the occupancy value $s_j$ of $\mathbf{p}_j$,
\begin{equation}
f_s(\mathbf{p}_j, \mathbf{S}_{1,j}^p,\mathbf{S}_{2,j}^p, \dots, \mathbf{S}_{n,j}^p) := s_j \ .
\end{equation}
The completed mesh structure $\hat{\mathbf{X}}$ is obtained by applying marching cubes~\cite{lorensen1987marching} on the voxel grid $\mathbf{\mathcal{X}}_s$ with the predicted occupancy values $s_j$.

\vspace{0.1cm}
\noindent \textbf{Texture Completion:} For the texture completion, the colored voxel grid $\mathbf{\mathcal{X}}_c$ is encoded using a 3D convolutional encoder $g_c$ to obtain a set of $m$ multiscale texture features,
\begin{equation}
g_c(\mathbf{\mathcal{X}}_c) := \mathbf{C}_1, \mathbf{C}_2, \dots, \mathbf{C}_m  \ ,
\end{equation}
where $\mathbf{C}_i \in \mathbb{R}^{r_i\times L \times L \times L}$, $1 \leq i \leq m$,  denote the texture features with resolution $L=\frac{N}{2^{i-1}}$ and channel dimension $r_i=r_1 \times 2^{i-1}$. Texture completion is carried out by predicting the RGB values of the query points $\{\mathbf{q}_j\}_{j=1}^M$, where $\mathbf{q}_j \in \mathbb{R}^3$. The points $\mathbf{q}_j$ are sampled from the ground-truth $\mathbf{Y}$ during training. At the inference time, $\mathbf{q}_j$ are the vertices of the reconstructed mesh $\hat{\mathbf{X}}$. Similar to the shape completion, the grid sampling function $\phi(\mathbf{C}_i, \mathbf{q}_j)$ allows extracting multiscale texture features $\mathbf{C}_{i,j}^q$ for each point $\mathbf{q}_j$. 

% The point encoding of $\mathbf{q}_j$ is denoted as $\phi(\mathbf{T}_i, \mathbf{q}_j)$ and
% \begin{equation}
% \{\phi(\mathbf{T}_i, \mathbf{q}_j)\}_{i=1}^{h} = \{\phi(\mathbf{C}_i, \mathbf{q}_j)\}_{i=1}^m \cup \{\phi(\mathbf{S}_i, \mathbf{q}_j)\}_{i=1}^n.
% \end{equation}

%\vspace{0.15cm}
The shape and texture completions are fused at the texture decoder level. This is achieved by concatenating the multiscale shape and texture features before feeding them to a decoder $f_c$ consisting of sequential 1D convolutional layers. In particular, the RGB values $\mathbf{c}_j$ for each point $\mathbf{q}_j$ are given by, 
%The texture decoder $f_c$ uses both the shape and texture features

%Then, $\mathbf{q}_j$ and its point encoding $\phi(\mathbf{T}_i, \mathbf{q}_j)$ are decoded with $f_c$ consisting sequential 1D convolutional layers to predict the RGB values $\mathbf{c}_j$ of $\mathbf{q}_j$,
\begin{equation}
f_c(\mathbf{q}_j, \mathbf{S}_{1,j}^q,\mathbf{S}_{2,j}^q, \dots, \mathbf{S}_{n,j}^q, \mathbf{C}_{1,j}^q,\mathbf{C}_{2,j}^q, \dots, \mathbf{C}_{m,j}^q) := \mathbf{c}_j \ .
\end{equation}
Finally, vertex textures are obtained by attaching the predicted $\mathbf{c}_j$ to $\hat{\mathbf{X}}$.

\subsection{Texture Refinement}\label{subsec:Texture_Refinement}
%\textcolor{red}{[Anis:] }
The proposed joint shape and texture networks are able to predict the vertex textures of a completed mesh. However, the predicted vertex textures have two limitations: (1) they are predicted for the full body, including the existing regions of the partial body. Consequently, high-level texture details of input partial body might be lost in the predicted textured body mesh; (2) the resolution of the vertex textures depends on the resolution of the reconstructed shape. This implies that high-resolution vertex textures come at a cost of high-resolution predicted shape, which is not straightforward to obtain due to memory constraints. 

\begin{figure}[!ht]
    \centering
	\includegraphics[width=0.99\textwidth]{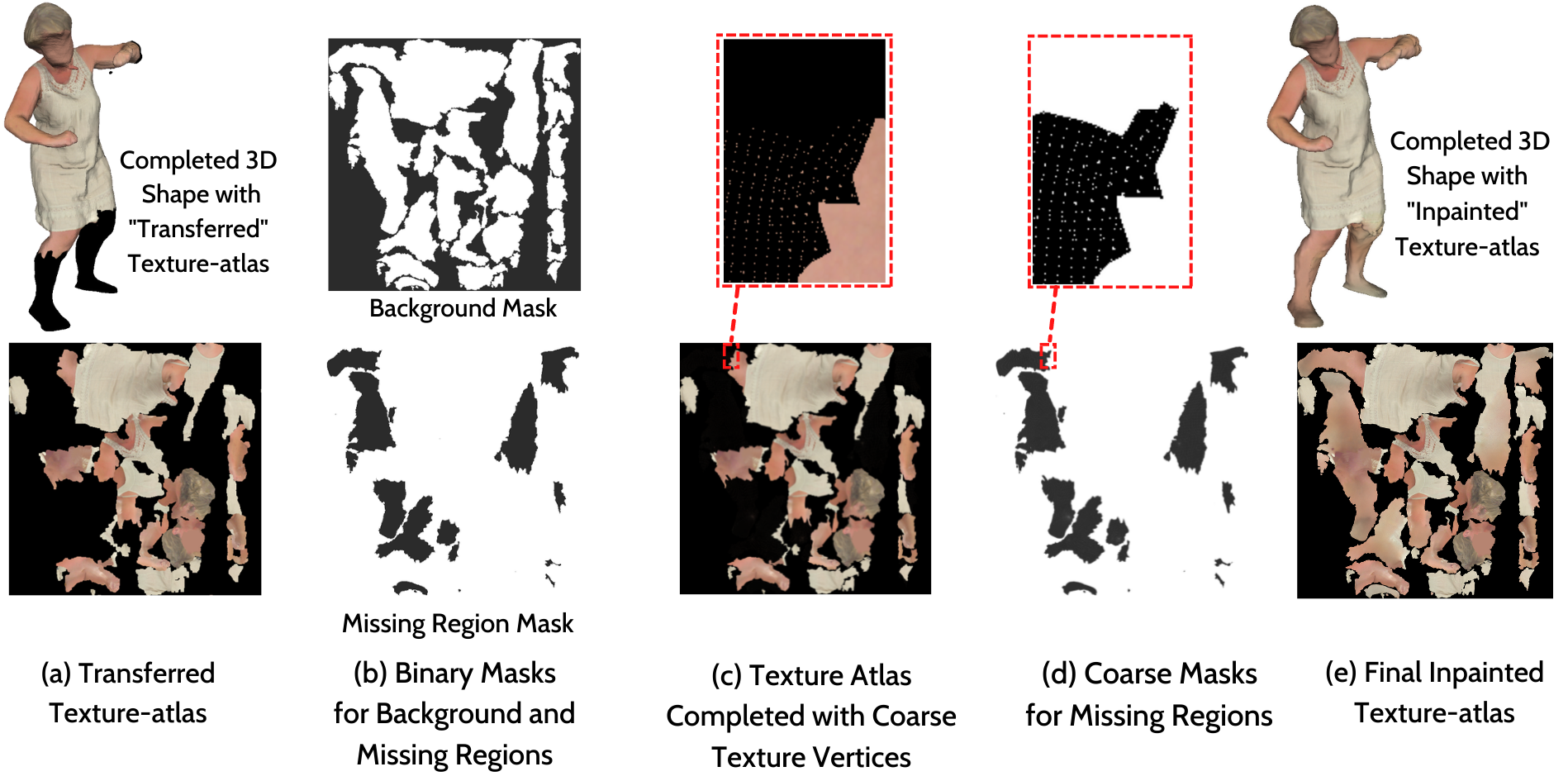}
    \caption{\textbf{Texture Refinement.} (a) the texture-atlas is transferred to the completed 3D shape, (b) the masks for the missing regions and background are identified, (c) the vertex textures are projected into the transferred texture-atlas, (d) the masks for missing regions are updated by unmasking the regions of the projected vertex textures, (e) the final inpainted texture-atlas.}
    \label{fig:Bcomnet_texture}
\end{figure}

%\vspace{0.1cm}
To overcome these issues, we use a \emph{texture-atlas}~\cite{levy2002least,saint20203dbooster} based refinement of the predicted vertex textures. This refinement reuses the coarse vertex textures predicted by the joint implicit shape and texture network and refines them in the 2D image space, while preserving the original texture from the partial input scan. In particular, the original texture of the partial scan is transferred to the completed shape by a ray-casting algorithm, as in~\cite{saint20203dbooster}. This allows the creation of a UV map and a texture atlas $\mathcal{A}$ for the completed mesh as depicted in Fig.~\ref{fig:Bcomnet_texture}(\textcolor{red}{a}). Following~\cite{saint20203dbooster}, the missing regions and the background regions are identified in the transferred texture atlas to create the two binary masks $M$ and $M_b$ shown in Fig.~\ref{fig:Bcomnet_texture}(\textcolor{red}{b}). Using the UV map created by the texture transfer, the vertex textures are projected to obtain a coarsely completed texture-atlas $\mathcal{A}_c$ as sketched in Fig.~\ref{fig:Bcomnet_texture}(\textcolor{red}{c}). The mask for missing regions $M$ is then updated with the projected vertex textures, yielding a coarse mask $M_c$ as displayed in Fig.~\ref{fig:Bcomnet_texture}(\textcolor{red}{d}).

%\vspace{0.1cm}
Given the coarsely completed texture-atlas $\mathcal{A}_c$, the coarse mask of missing regions $M_c$, and the background mask $M_b$, the problem of texture refinement is formulated as an image inpainting one. Specifically, we opt for the texture-atlas inpainting method proposed in~\cite{saint20203dbooster}, which adapts partial convolution inpainting~\cite{liu2018partialinpainting} to the context of texture-atlas. Partial convolutions in~\cite{liu2018partialinpainting,saint20203dbooster} extend standard convolutions
to convolve the information from unmasked regions (\emph{i.e.} white regions in the binary masks). 
Formally, let us consider a convolution filter defined by the weights $w$, the bias $b$ and the feature values $\mathcal{A}_c^{w}$ of the texture-atlas $\mathcal{A}_c$ for the current sliding window. Given the coarse mask of missing regions $M_c$ and the corresponding background mask $M_b$,  the partial convolution at every location, similarly defined in~\cite{saint20203dbooster}, is expressed as, 

% \begin{equation}
% t_c =
% \begin{cases}
%   W^T (T_w \odot M)
%   \cdot
%   \frac{\text{sum}(\mathbf{1})}{\text{sum}(M)} + b
%   &
%   \text{if}\; \text{sum}(M) > 0
%   \\
%   0
%   &
%   \text{otherwise}
%   \end{cases}
% \end{equation}

%One important observation in the two texture atlases provided in Fig.~\ref{fig:texture_atlas}, is that they contain some non-informative black regions used as background to gather the body charts in a single image.
%The inpainting of the missing texture information (white regions in Fig.~\ref{fig:identified_missing_partial}) could be impacted by the non-informative background (\ie black) using the original form of partial convolutions introduced in~\cite{liu2018partialinpainting}.
%This is confirmed and visualised by experiments in Section~\ref{sect:exp_inpaint}.
%As a solution, we propose to ignore these regions during the partial convolutions as done with the masked values of the missing texture to be recovered.
%However, these regions should not be updated during the mask update as the background mask should stay fixed
%through all the partial convolution layers.
%his is achieved by including the \emph{background mask} $M_b$ of the texture image
%in the partial convolution as follows, 

\begin{equation}
a_c =
\begin{cases}
  w^T (\mathcal{A}_c^{w} \odot M_c \odot M_b)
  \cdot
  \frac{\text{sum}(\mathbf{1})}{\text{sum}(M_c \odot M_b)} + b
  & \text{if}\; \text{sum}(M_c \odot M_b) > 0
  \\
  0
  & \text{otherwise}
\end{cases} \ , 
\end{equation}

\noindent where $\odot$ denotes element-wise multiplication, and $\mathbf{1}$ has same shape as $M_c$ but with all elements being $1$. As proposed in~\cite{saint20203dbooster}, the masks $M_c$ are updated after every partial convolution, while the background masks are passed to all partial convolutions layers without being updated by applying $\emph{do-nothing}$ convolution kernels. The partial convolutional layers are employed in a UNet architecture~\cite{ronneberger2015u} instead of standard convolutions. At training time, the vertex textures are sampled from the ground-truth texture-atlas. The same loss functions in~\cite{liu2018partialinpainting,saint20203dbooster} are used to train the network. 
\vspace{0.1cm}
It is important to highlight that the proposed texture refinement is different from~\cite{saint20203dbooster} as it reuses the predicted vertex textures instead of inpainting the texture-atlas from scratch. We show in Section~\ref{sec:Experiments_Results_Evaluation} that the proposed refinement outperforms the inpainting from scratch and the vertex texture based completion. Furthermore, we reveal that such refinement can be used to improve other vertex based texture completion. 
%a masked value in $M$ ($M(i,j)=0$) is updated to unmasked ($M(i,j)=1$) if the convolution was able to condition its output on at least one valid input value. In practice this is achieved by applying fixed convolutions, with the same kernel size as the partial convolution operation, but with weights identically set to 1 and no bias. The background mask $M_b$ is passed to all partial convolutions layers without being updated by applying $\emph{do-nothing}$ convolution kernels with the same shape as the ones used for the masks $M$. A $\emph{do-nothing}$ kernel consists of a kernel with zeros values everywhere except for the central location which is set to 1. Moreover, before updating the original mask $M$ we apply on it this background mask $M_b$ by element-wise multiplication so that we guarantee that the mask $M$ will not be updated using the background regions. 

%Several loss functions are used to optimize the network.
%Two pixel-wise reconstruction losses are defined separately on the masked and unmasked regions with a
%focus on masked regions.
%Style transfer losses are also considered by constraining the feature maps of the predictions and their auto-correlations to be close those of the ground truth~\cite{gatys2015neural}.

\section{Experimental Results}\label{sec:Experiments_Results_Evaluation}
\noindent\textbf{Dataset and Evaluation Metrics.}
% \subsection{Dataset and Evaluation Metrics.}
Our method was trained on the 3DBodyTex.v2 dataset~\cite{saint20183dbodytex,saint2019bodyfitr,saint2020sharp} which has been recently used as a benchmark for the SHARP challenge~\cite{SHARP2022}. The dataset contains a large variety of human poses and different clothing types from $500$ different subjects. Each subject is captured with $3$ different poses and different clothing types, such as close-fitting or arbitrary casual clothing. The number of ground truth scans in the training set is $2094$. A number of $15904$ partial scans for training and validation were generated using the routines provided by the SHARP challenge organizers~\cite{SHARP2022Repo}. The number of unseen (during training) scans in the evaluation set is $451$. 
%The dataset also contains different subsets T1 and T2 with different types of partiality where T2 is more challenging with respect to the amount of available partiality.

As considered in SHARP challenges, the evaluation is conducted in terms of shape scores $S_s$, texture scores $S_t$, and the area scores $S_a$. Shape and texture scores are calculated via measuring surface-to-surface distances by sampling points from the ground truth and the reconstructed meshes. The area score estimates the similarity between the triangle areas of the meshes. The final score $S_r$ is calculated as $S_r = \frac{1}{2}S_a(S_s + S_t)$. Details about these metrics can be found in~\cite{saint2020sharp,SHARP2022Repo}\footnote{The details about the metrics can be accessed via: \url{https://gitlab.uni.lu/cvi2/cvpr2022-sharp-workshop/-/blob/master/doc/evaluation.md}}. The evaluation of the completed meshes is performed via the Codalab system provided by the SHARP challenge\footnote{The leaderboard on Codalab can be accessed via: \url{https://codalab.lisn.upsaclay.fr/competitions/4604\#results}}.

% It is important to mention that the reported scores are the result of mapping distance values (shape or texture) to scores in percentage using a parametric function\footnote{The original parameters of SHARP challenge 2022 metrics are used.}. This mapping might make the scores from different methods close to each other, depending on the chosen parameters.

\vspace{-0.4cm}
\subsection{Network Training Details}
We trained the joint implicit networks using the Adam optimizer with a learning rate of $10^{-4}$ for $54$ epochs. The model was trained on an NVIDIA RTX A6000 GPU using the Pytorch library. The query points for the training are obtained from the ground truth surfaces by sampling $100000$ points. During training, we sub-sample $50000$ of these points at each iteration. Gaussian random noise $\mathcal{N(\sigma)}$ is added to each point to move the sampled point near or far from the surface, depending on the $\sigma$ value. Similar to \cite{chibane2020ifnet_texture}, the $\sigma$ is chosen as $0.01$ for half of the points and $0.1$ for the other half. We did not add noise to the query points of the texture network, as its goal is to predict the color value of the points sampled from the surface. Partial scans are voxelized by sampling $100000$ points from the partial surface and setting the occupancy value in the nearest voxel grid to $1$. Similarly for the colored voxelization, the value of the nearest voxel is set to the RGB value of the sampled point obtained from the corresponding texture-atlas. The input voxel resolution is $128$ and the resolution for the final retrieval is $256$.

%\vspace{0.1cm}
%T2 subset of the SHARP dataset was used for training. 
We follow a similar naming convention as in~\cite{wang2018high} for our architecture details. Let \textit{c3-k} denote Conv3D-ReLU-BatchNorm block and \textit{d3-k} denote two Conv3D-ReLU blocks followed by one BatchNorm layer where \textit{k} is the number of filters. The kernel size for all 3D convolutional layers is $3\times3$. Let \textit{gs-i} represent the grid feature sampling of the query points from the output of the previous layer, and \textit{mp} denote 3D max pooling layer. Let \textit{c1-k} denote the Conv1D layer with 1x1 kernel where $k$ is the number of output features and ReLU activation except the output layer where the activation is linear. The encoder architecture for shape and texture networks is composed of the following layers: \textit{gs-0, c3-16, gs-1, mp, d3-32, d3-32, gs-2, mp, d3-64, d3-64, gs-3, mp, d3-128, gs-4, mp, d3-128, d3-128, gs-5}.\footnote{Sampling features \textit{gs-1, gs-2, ..., gs-5} are flattened and concatenated. Shape features are also added here for the texture encoding.} The decoder architecture for the shape network contains: \textit{c1-512, c1-256, c1-256, c1-1}. The decoder architecture for the texture network consists of \textit{c1-512, c1-256, c1-256, c1-3}. The partial convolutional network is trained with Adam optimizer and learning rate of $10^{-4}$ for 330000 iterations. The original texture size of $2048\times2048$ was used for training the network with a batch size of $1$.
%\vspace{0.1cm}

%

%
%
\subsection{Results and Evaluation}
In this section, a qualitative and quantitative analysis of the results of TSCom-Net against the results of the SHARP challenge participants are presented. Other participants of the challenge employ implicit networks for the shape reconstruction with vertex-based texture completion~\cite{SHARP2022}. We also compare our method to the state-of-the-art IFNet-Texture~\cite{chibane2020ifnet_texture} method, which won the previous editions of SHARP 2021 and 2020~\cite{saint2020sharp}.  The shape network of~\cite{chibane2020ifnet_texture} is re-trained with the generated partial data of SHARP 2022. For the vertex texture predictions, the available pretrained model provided by the authors was used.

\vspace{0.15cm}
\noindent\textbf{Quantitative Evaluation:} In Table~\ref{tab:results}, we illustrate the quantitative results using the metrics of SHARP challenge~\cite{saint2020sharp}. Overall, our method is ranked second in the leaderboard with a final score of $84.73\%$, obtaining a better texture score than Method-Rayee, Method-Raywit and IFNet-Texture~\cite{chibane2020ifnet_texture}. It should be noted that the texture score depends on the shape score as well, since it is not possible to get a correct texture score for an incorrectly predicted shape.
\vspace{-0.3cm}
\begin{table}[t]
\centering
\begin{tabular}{@{} l | c c c | c}
\hline
    \thead{\textbf{Method}}
    & 
    \thead{\textbf{Shape}\\ \textbf{Score(\%)} }
    & 
    \thead{\textbf{Area}\\ \textbf{Score(\%)}}
    & 
    \thead{\textbf{Texture}\\ \textbf{Score(\%)}}
    & 
    \thead{\textbf{Final}\\ \textbf{Score(\%)}}\\
    \hline
    \hline
    IFNet-Texture\cite{chibane2020ifnet_texture} &  85.44 $\pm$ 2.93 & 96.26 $\pm$ 6.35& 81.25 $\pm$ 7.61 & 83.34 $\pm$ 6.86
    \\
    Method Raywit & 85.91 $\pm$ 7.14& 93.96 $\pm$ 3.96& 83.45 $\pm$ 8.43& 84.68 $\pm$ 7.63\\
    
    Method Rayee & \textcolor{gray}{ \textbf{86.13 $\pm$ 7.32}} & 96.26 $\pm$ 3.61 & 83.23 $\pm$ 8.31 & 84.68 $\pm$ 7.74
    \\
    Method Janaldo & \textbf{89.76 $\pm$ 4.97} & \textbf{96.76 $\pm$ 2.28} & \textbf{87.10 $\pm$ 6.33} & \textbf{88.43 $\pm$ 5.56}\\
    
    % TSCom-Net (\textbf{Ours}) & \textcolor{gray}{ \textbf{85.75}} & \textcolor{gray}{ \textbf{96.68}} & \textcolor{gray}{ \textbf{83.74}} & \textcolor{gray}{ \textbf{84.75}}\\
    TSCom-Net (\textbf{Ours}) &  85.75 $\pm$ 6.15 & \textcolor{gray}{ \textbf{96.68$\pm$ 2.89}} & \textcolor{gray}{ \textbf{83.72 $\pm$ 6.95}} & \textcolor{gray}{ \textbf{84.73 $\pm$ 6.5}}\\
    \hline
\end{tabular}
\vspace{0.15cm}
\caption{Quantitative Results for SHARP 2022. The best and second best scores are denoted in \textbf{bold-black} and \textcolor{gray}{\textbf{bold-gray}} colors respectively.}
\label{tab:results}
\end{table}
\begin{table}[!ht]
\centering
\begin{tabular}{@{} l | c}
\hline
    \thead{\textbf{Method}}
    & 
    \thead{\textbf{Texture Score (\%)}}\\
     %\thead{\textbf{\%}}\\
    \hline
    \hline
    Method-Janaldo & 87.10$ \pm$ 6.33\\
    
    Method-Janaldo + Our Texture Refinement & \textbf{87.54 $\pm$ 6.19}\\
    \hline
\end{tabular}
\vspace{0.15cm}
\caption{Effectiveness of Our Texture Refinement.}
\label{tab:compare_janaldo}
\end{table}

% \vspace{0.1cm}
Furthermore, we demonstrate the effectiveness of our texture refinement method by applying it to the predictions of Method-Janaldo. The results in Table~\ref{tab:compare_janaldo} show that our inpainting network introduced in Section~\ref{subsec:Texture_Refinement} improves the texture score by $0.44\%$. This shows that the proposed texture refinement can be used as an additional component to improve other shape and texture completion methods that predict low-resolution vertex textures. 
%we also compared using our texture-refinement network on the results of Method-Janaldo with their original results. Table~\ref{tab:compare_janaldo} shows the results of refining the texture of Janaldo with our inpainting network where we improve the texture score by 0.44\%.
%
%
%
\begin{figure}[!ht]
    \centering
	\includegraphics[width=0.99\textwidth]{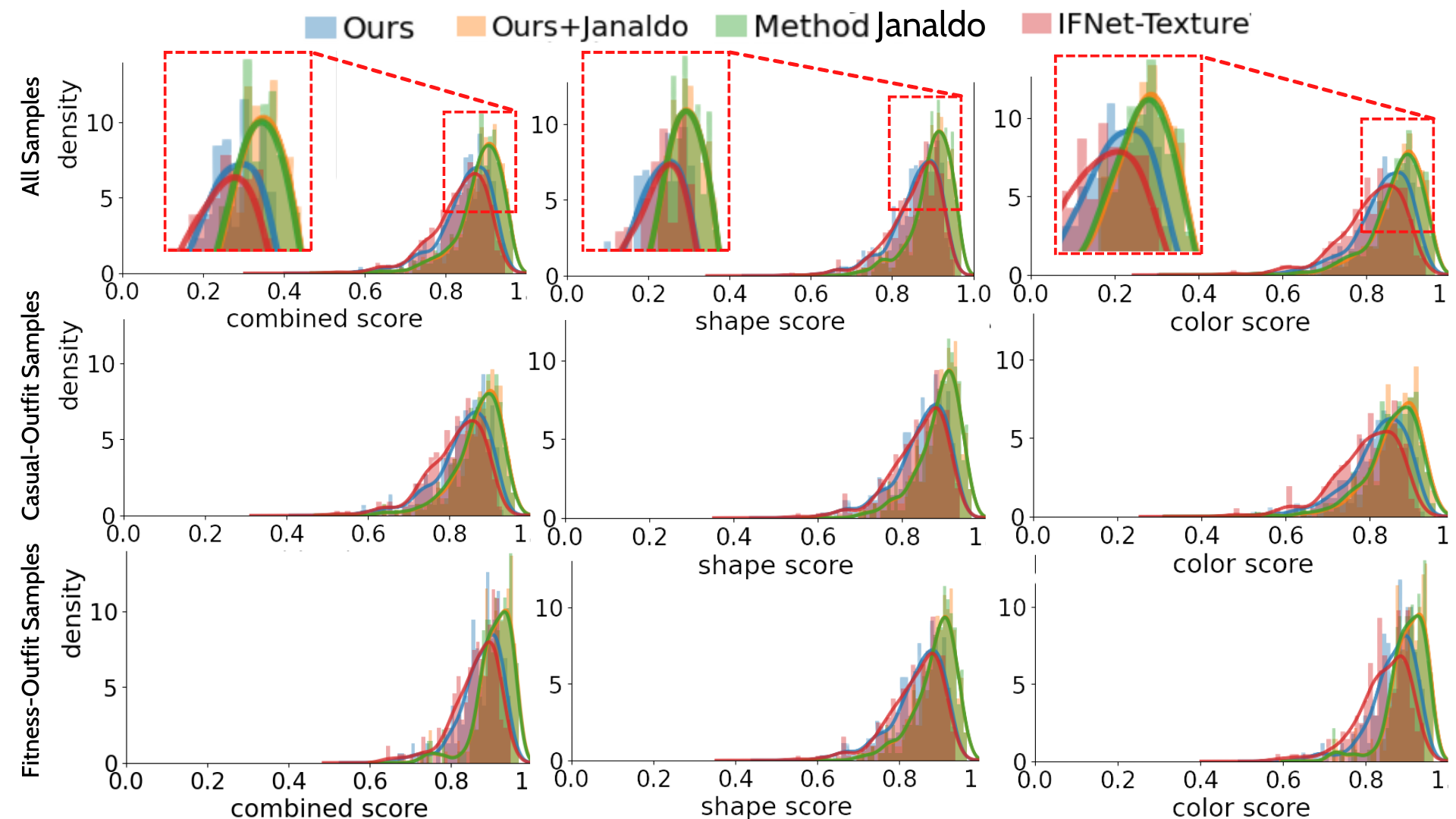}
    \caption{\textbf{Score Distribution. } Distribution of shape, color and combined final scores ($\frac{\text{score}}{100}$ along x-axes of the plots). The first row reports the distribution of the scores on all the scans. The second row focuses on casual-outfit scans while the third one reports the distributions on fitness-outfit scans.} 
    \label{fig:Bcomnet_EvalPlot}
\end{figure}

\vspace{0.1cm}
Finally, Fig.~\ref{fig:Bcomnet_EvalPlot} show the distributions of shape, texture, and final scores for the predictions of TSCom-Net, Method-Janaldo, Method-Janaldo with our texture refinement, and IFNet-Texture~\cite{chibane2020ifnet_texture}. The first row of this figure illustrates the overall distribution of the scores and highlights the superiority of our texture scores \textit{w.r.t.}~IFNet-Texture~\cite{chibane2020ifnet_texture}. While Method-Janaldo outperforms our results, it can be observed that if we endow it with our texture refinement scheme, the distribution of texture scores becomes slightly higher. In the second and third row of Fig.~\ref{fig:Bcomnet_EvalPlot}, we report the distribution of the scores on scans with casual-outfit and fitness-outfit. Unsurprisingly, casual-outfit scans were more challenging than fitness-outfit ones for all methods. Nevertheless, our approaches recorded more improvements \textit{w.r.t.}~Method-Janaldo and IFNet-Texture on the casual-outfit scans than the fitness-outfit ones.

\begin{figure}[!h]
    \centering
	\includegraphics[width=0.9\textwidth]{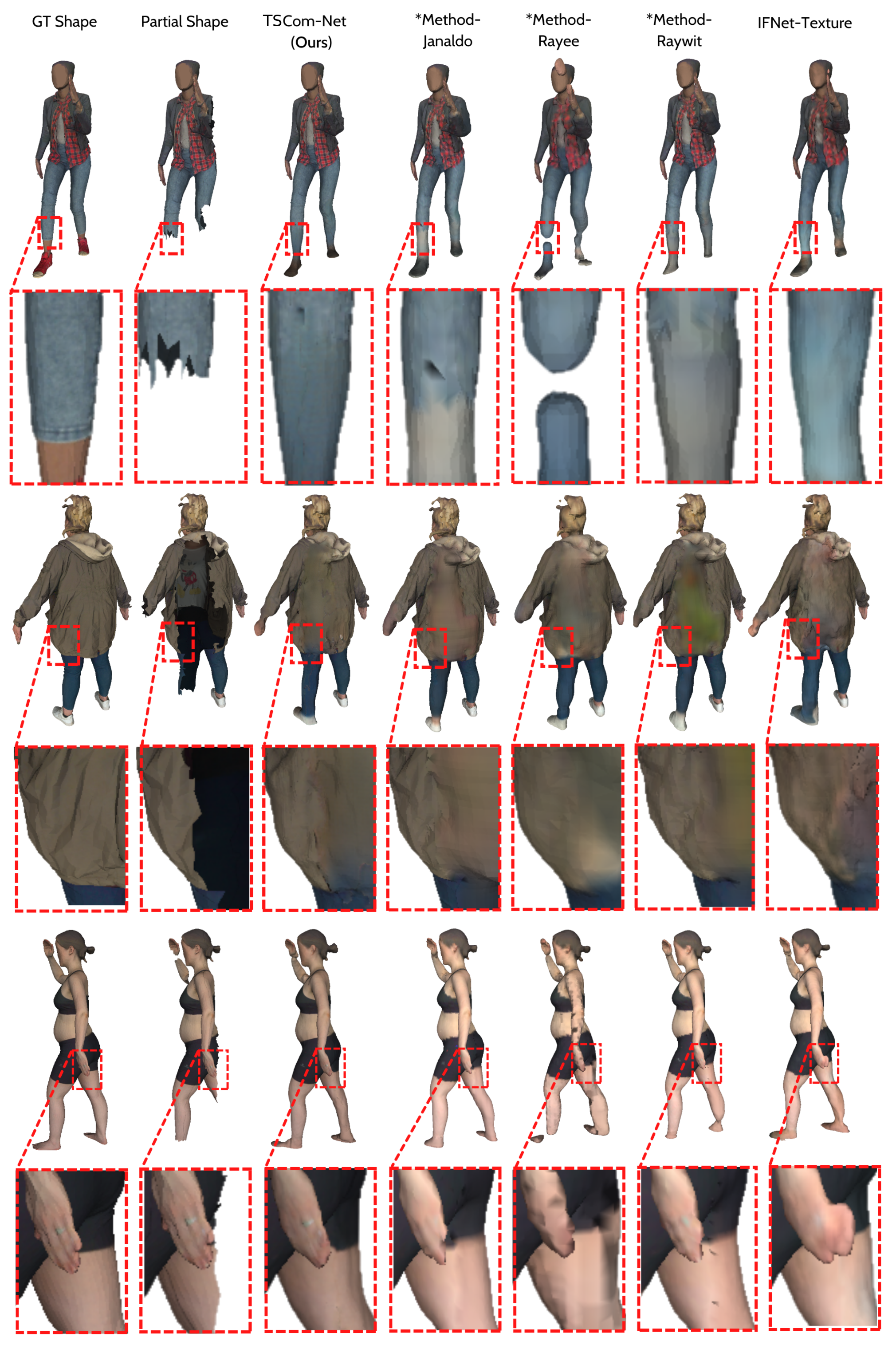}
    \caption{\textbf{Qualitative results. } Visual comparisons of textured body shape completion results by different competing methods (as per Table.\ref{tab:results}). The \textit{first two rows} depict results on samples with casual outfits, \ie more variation on garment texture pattern. The \textit{last two rows} depict results on samples with fitness outfits.
    }
    \label{fig:ExpResults1}
\end{figure}

It is important to mention that the reported scores are the result of mapping distance values (shape or texture) to scores in percentage using a parametric function. This mapping might make the scores from different methods close to each other, depending on the chosen parameters. We note that the original parameters of the SHARP challenge 2022 metrics are used. Thus, we further conducted a qualitative evaluation of our method.

\vspace{0.15cm}
\noindent\textbf{Qualitative Evaluation:}
% \textcolor{red}{transition metric} 
Fig.~\ref{fig:ExpResults1} shows a qualitative comparison of our approach to other competing methods on models with casual outfit and fitness outfit. Considering the coat missing region of the top row model, it can be noted that IFNet-Texture~\cite{chibane2020ifnet_texture} and Method-Raywit are unable to predict the correct colors. On the other hand, Method-Rayee and Method-Janaldo produce over-smoothed textures, creating blurry artifacts. Neither of these effects can be observed on the TSCom-Net predictions. 
In the second row, the bottom legs and feet appear to be the most difficult regions to recover. Method-Rayee and Method-Raywit fail to produce the correct shape for these missing regions. While IFNet-Texture~\cite{chibane2020ifnet_texture} and Method-Janaldo are able to recover the correct shape, both generate white color artifacts on the jeans. Our method is the only one to produce more reasonable texture predictions, showing a sharper change in color between the jeans and the feet.
Similar to the second model, it is apparent that our results on the third model are sharper for the regions with a color change from skin to black when compared to the other results. 
%As for the previous models, TSCom-Net appears to produce the most plausible prediction for the bottom row model when considering both the shape and texture.
% \textcolor{red}{discussion metric}
The visual comparisons illustrate that our results are of higher resolution with better texture representation than the competing approaches despite the close quantitative scores.

\subsection{Ablation Study}\label{subsec:Ablation_Study}
\vspace{-0.5cm}
\begin{figure}[!ht]
    \centering
	\includegraphics[width=0.99\textwidth]{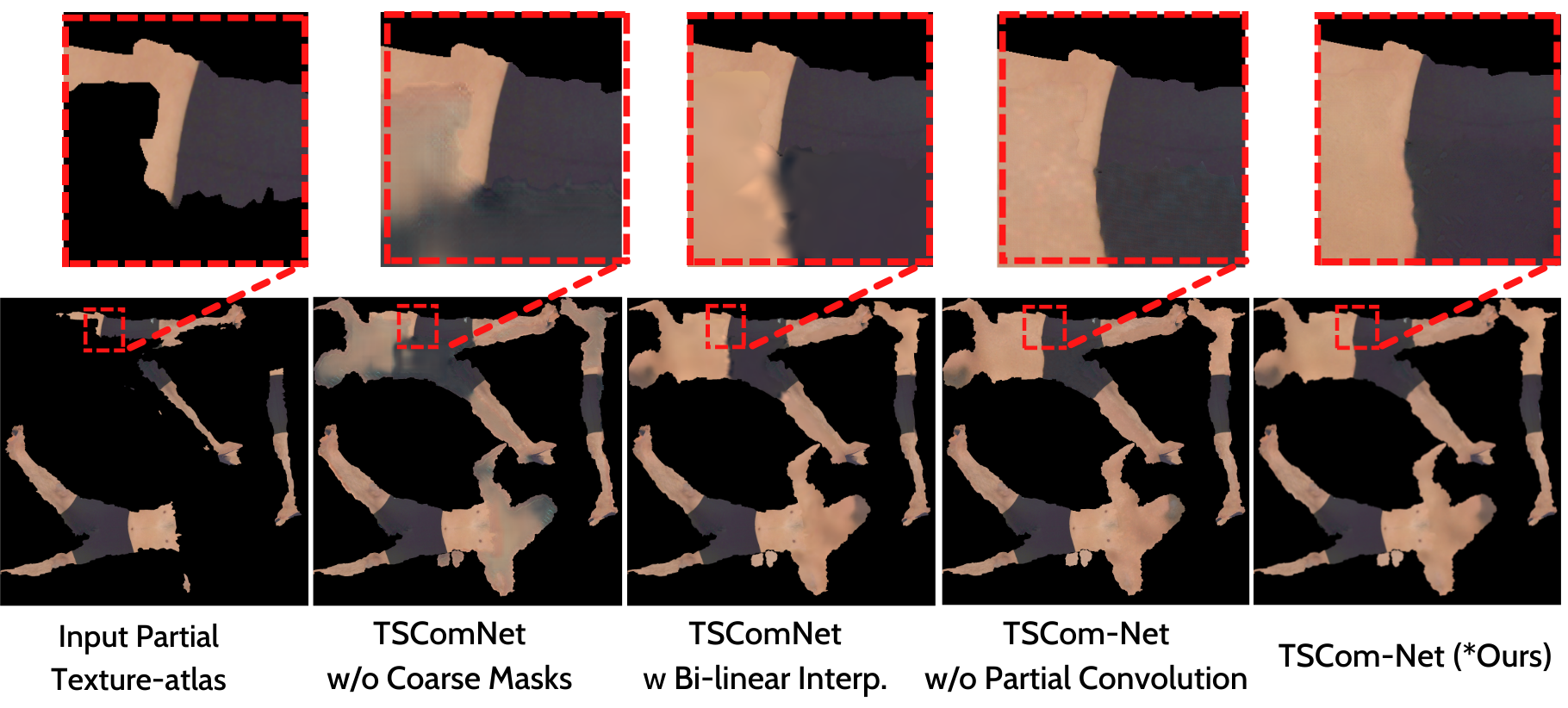}
	\vspace{-0.3cm}
    \caption{Multi-stage improvements of texture-atlas inpainting in TSCom-Net.} 
    \label{fig:Texture_Inpaintg_Artifacts}
\end{figure}
\vspace{-1cm}
\begin{table}[ht]
\centering
\begin{tabular}{@{} l | c c c | c}
    \hline
    
    \thead{\textbf{Method}}
    & 
    \thead{\textbf{Shape}\\ \textbf{Score(\%)}}
    & 
    \thead{\textbf{Area}\\ \textbf{Score(\%)}}
    & 
    \thead{\textbf{Texture}\\ \textbf{Score(\%)}}
    & 
    \thead{\textbf{Final}\\ \textbf{Score(\%)}}\\
    \hline
    \hline
    IFNet-Texture~\cite{chibane2020ifnet_texture} & 85.44 $\pm$ 2.93 & 96.26 $\pm$ 6.35 & 81.25 $\pm$ 7.61 & 83.34 $\pm$ 6.86\\
    
    Texture-transfer Baseline & 85.75$ \pm$ 6.15 & 96.68 $\pm$ 2.89 & 56.51 $\pm$ 18.98 & 71.13 $\pm$ 11.11 \\
    
    Ours w/o Coarse Masks & 85.75$ \pm$ 6.15 & 96.68 $\pm$ 2.89  & 81.04 $\pm$ 7.92 & 83.39$ \pm$ 6.87\\
    
    Ours w/o Tex. Refine. & 85.75$ \pm$ 6.15 & 96.68 $\pm$ 2.89  & 83.27 $\pm$ 7.08 & 84.53 $\pm$ 6.54 \\
    
    Ours w/ Bilinear Interp. & 85.75 $\pm$ 6.15 & 96.68  $\pm$ 2.89 & 83.66 $\pm$ 6.95 & 84.71 $\pm$ 6.50 \\

    Ours w/o Partial Conv. & 85.75 $\pm$ 6.15 & 96.68  $\pm$ 2.89 & 83.68 $\pm$ 6.96 & 84.71 $\pm$ 6.50\\
    
    \textbf{TSCom-Net (Ours)} &  \textbf{85.75 $\pm$ 6.15} & \textbf{96.68$\pm$ 2.89} & \textbf{83.72 $\pm$ 6.95} & \textbf{84.73 $\pm$ 6.50}\\
    %\textbf{TSCom-Net (Ours)} &  \textbf{85.75 $\pm$ 6.15} & \textbf{96.68$\pm$ 2.89} & \textbf{83.72 $\pm$ 6.95} & \textbf{84.73 $\pm$ 6.5}\\
    \hline
\end{tabular}
\vspace{0.1cm}
\caption{Quantitative scores of TSCom-Net with multi-stage improvements.}
\label{tab:ablation}
\end{table}
\vspace{-0.5cm}
\noindent Our joint implicit function learning for shape and texture (TSCom-Net w/o Texture Refinement) gives a $1.19\%$ (cf. Table \ref{tab:ablation}) increase in the final score with respect to~\cite{chibane2020ifnet_texture} demonstrating the effect of early fusion between SCom-Net and TCom-Net. Only transferring the partial texture to the reconstructed shape, where the color values for the missing regions are all black, gives us a baseline texture score of $56.51\%$. Training an inpainting network directly on the partial textures (TSCom-Net w/o Coarse Masks) increase texture score to $81.04\%$ which is $0.21\%$ lower than \cite{chibane2020ifnet_texture} and $24.53\%$ higher than the baseline. Conducting bilinear interpolation of the vertex colors in missing regions is giving a texture score of $83.66\%$. 
% which is higher than TSCom-Net with vertex colors (TSCom-Net w/o Texture Refinement). 
All-in-all partial convolutions, instead of standard convolutions, gives stable and more sharper texture inpainting results. Fig. \ref{fig:Texture_Inpaintg_Artifacts} depicts how the different types artifacts or lack of sharpness appear when other options of inpainting are tested. Finally, TSCom-Net consisting of all the components is giving the highest final score of $84.73\%$.
\begin{table}[ht]
\centering
\begin{tabular}{l|cc|cc|c}
\hline
\multicolumn{1}{c|}{\multirow{2}{*}{\textbf{Method}}} 
& \multicolumn{2}{c|}{\textbf{Partiality Type}}    
& \multirow{2}{*}{\thead{\textbf{Shape}\\ \textbf{Score (\%)}}}  
& \multirow{2}{*}{\thead{\textbf{Texture}\\ \textbf{Score (\%)}}} 
& \multirow{2}{*}{\thead{\textbf{Final}\\ \textbf{Score (\%)}}} \\
\vspace{0.1cm}
%\multicolumn{1}{c}{} 
& Training
% \multicolumn{1}{c}{} 
% & \multicolumn{1}{c|}{Training} 
& Testing 
&
&
&\\ 
\hline
IFNet-Texture~\cite{chibane2020ifnet_texture} 
        & \multicolumn{1}{c|}{T2}  & T1  & 87.99   $\pm$ 4.65                       & 84.32   $\pm$ 5.73                       & 86.15  $\pm$ 5.09                      \\
IFNet-Texture~\cite{chibane2020ifnet_texture}            & \multicolumn{1}{c|}{T1}  & T1   & 86.49 $\pm$ 3.96                       & \textbf{89.28 $\pm$ 2.89}                         & \textbf{87.88 $\pm$ 3.32}                        \\
3DBooster~\cite{saint20203dbooster}                                    & \multicolumn{1}{c|}{T1}       & T1      & 58.81    $\pm$ 14.99                    & 72.31 $\pm$ 6.79                          & 65.57  $\pm$ 3.32       \\
TSCom-Net (\textbf{Ours})                                         & \multicolumn{1}{c|}{T2}       & T1      & \textbf{88.05 $\pm$ 4.84}                        & 85.46  $\pm$  5.56 & 86.75 $\pm$  5.14                      \\\hline              
\end{tabular}
\vspace{0.1cm}
\caption{Generalizability of TSCom-Net when trained and tested on samples with different types of partiality.}
\label{tab:t1_generalization}
\end{table}
\vspace{-0.3cm}
\noindent\textbf{{Generalization to Other Types of Partiality:} }
In this section, we evaluate the generalization capability of our model to other types of partiality. In particular, we consider the \textit{view-based} partiality (T2) introduced in SHARP 2022~\cite{SHARP2022} to train the models and use the \textit{hole-based} partiality (T1) introduced in previous editions of SHARP~\cite{saint2020sharp} for inference. This implies that the networks trained on (T2) have never seen hole-based partial scans (T1). Table~\ref{tab:t1_generalization} demonstrates a comparison of the generalization capability of our method to IFNet-Texture~\cite{chibane2020ifnet_texture} that is also trained on (T2) and tested on (T1) subset. Following these settings, we obtain an increase of $1.14\%$ for the texture score compared to \cite{chibane2020ifnet_texture}. We also compare our model trained on (T2) and tested on (T1) to IFNet-Texture~\cite{chibane2020ifnet_texture} and 3DBooster~\cite{saint20203dbooster} both trained on (T1) and tested on (T1). In this case, our approach significantly outperforms~\cite{saint20203dbooster} while achieving comparable results to~\cite{chibane2020ifnet_texture} although both were trained on (T1).

\vspace{-0.1cm}
\section{Conclusion}\label{sec:Conclusion}
\vspace{-0.1cm}
This paper presents a method for completing the shape and texture of 3D partial scans. Joint implicit feature networks are proposed for learning to complete the shape and textures. Moreover, a new coarse-to-fine texture refinement network was introduced. It generates high-resolution texture from the predicted coarse vertex texture and the available partial texture. Experimental evaluations show that our method gives visually more appealing results than the state-of-the-art and is positioned second in the SHARP 2022 challenge. In future, we plan to make the entire TSCom-Net modules end-to-end trainable for completing 3D scans and refining the texture. At the same time, we will investigate neural implicit radiance field for texture completion (with editable 2D UV texture map) and 3D surface reconstruction. 
\clearpage
\bibliographystyle{splncs04}
\bibliography{egbib}
\end{document}